\documentclass[11pt]{article}

\usepackage[a4paper,margin=1in]{geometry}
\usepackage{amsmath,amssymb,amsfonts,mathtools}
\usepackage{bm}
\usepackage{microtype}
\usepackage{hyperref}
\usepackage{enumitem}
\usepackage{titlesec}
\usepackage{cite}
\usepackage{mathrsfs}
\usepackage{graphicx}
\usepackage{tikz}

\titleformat{\section}{\large\bfseries}{\thesection.}{0.5em}{}
\titleformat{\subsection}{\normalsize\bfseries}{\thesubsection.}{0.5em}{}

\title{\textbf{Diffusion Maps is not Dimensionality Reduction}}
\author{Julio Candanedo\footnote{SparseTrace.ai. Appleton, WI, USA. \url{julio@sparsetrace.ai}.} \qquad\qquad Alejandro Patiño\footnote{Universidad Nacional de Colombia sede Manizales. Manizales, Caldas, Colombia}}
\date{}

\newcommand{\diag}{\mathrm{diag}}

\begin{document}
\maketitle

\large

\begin{abstract}
Diffusion maps (DMAP) are often used as a dimensionality-reduction tool, but more precisely they provide a spectral representation of the intrinsic geometry rather than a complete charting method. To illustrate this distinction, we study a Swiss roll with known isometric coordinates and compare DMAP, Isomap, and UMAP across latent dimensions. For each representation, we fit an oracle affine readout to the ground-truth chart and measure reconstruction error. Isomap most efficiently recovers the low-dimensional chart, UMAP provides an intermediate tradeoff, and DMAP becomes accurate only after combining multiple diffusion modes. Thus the correct chart lies in the span of diffusion coordinates, but standard DMAP do not by themselves identify the appropriate combination. 
\end{abstract}

%\tableofcontents

\section{Introduction}

High-dimensional data are often modeled as samples from, or near, a low-dimensional Riemannian manifold $\mathscr{M}$, in the sense of the manifold hypothesis \cite{fefferman2016_testingManifoldHypothesis}. In this setting, the natural geometric object is not a coordinate chart chosen in advance, but the manifold itself together with its intrinsic differential operators, foremost the Laplace--Beltrami (LB) operator $\Delta_{\mathscr{M}}$. On a compact manifold, the eigenfunctions $\{\phi_n\}_{n\geq 0}$ of $\Delta_{\mathscr{M}}$ form an orthonormal basis of $L^2(\mathscr{M})$, so any sufficiently regular scalar function $f:\mathscr{M}\to\mathbb{R}$ admits a spectral expansion, a manifold-based Fourier-transform:
\begin{align}
    f(r) = \sum_{n\in\mathbb{N}} L_n \phi_n(r)\quad.
\end{align}
Applied coordinate-wise to an embedding $F:\mathscr{M}\to\mathbb{R}^D$, this yields a discrete manifold analogue of a discrete Fourier expansion: each ambient coordinate function $F^x$ can be expressed in the Laplace--Beltrami basis, at least in the infinite-mode limit \cite{gine2006_empiricalGraphLaplacian, jones2008_manifoldParametrizationEigenfunctions, bates2014_laplacianEigenfunctionEmbedding}. In sampled form, this suggests a discrete relation of the type
\begin{align}
    R_{ix} \approx \sum_n L_{nx}\phi_{in} ,
\end{align}
where $\phi_{in}$ denotes sampled LB eigenfunctions and $L_{nx}$ the corresponding spectral coefficients. From this perspective, the essential object is a spectral representation of geometry, not yet a low-dimensional embedding in the sense of a chart or compression map.

A discrete version of this construction is provided by graph-based Laplacian methods. Laplacian Eigenmaps showed that one can approximate manifold geometry by eigenvectors of a graph Laplacian constructed from sampled data, while also emphasizing that such coordinates do not in general furnish an isometric embedding of the manifold \cite{berard1994_heatKernelEmbedding, belkin2003_laplacianEigenmaps}. Diffusion-maps (DMAP) sharpen this picture by introducing a Markov normalization of the kernel graph together with the Coifman--Lafon drift correction \cite{coifman_2006_diffusion_maps, nadler2005_diffusionMapsFokkerPlanck}. In this sense, diffusion coordinates should first be read as eigenfunctions of a data-driven diffusion operator, with the Nyström extension supplying an interpolation rule away from the training set, rather than as immediate evidence of dimensionality reduction \cite{bengio2003_nystromOutOfSample}.

This distinction is the starting point of the present note. Methods such as Isomap (IMAP), UMAP, PHATE, or t-SNE are explicitly designed to produce a low-dimensional embedding \cite{tenenbaum2000_isomap, vandermaaten2008_tsne, mcinnes2018_umap, moon2019_phate, gildenblat2026_globalStructurePreservationDR}. DMAP by contrast, begin with an operator and its spectrum. A finite collection of diffusion modes may furnish an accurate approximation to diffusion distance, and in special cases may support a useful parametrization, but this is a downstream truncation of a richer spectral object rather than the defining act of the method itself.
 
To discretize the smooth manifold picture into an sampled-data construction, let the dataset matrix $R_{iX}\in\mathbb{R}^{N\times D}$ denote the ambient coordinates of $N$ observations, and define the pairwise squared Euclidean distances $D^2 = D^2_{ij} = \|R_{iX}-R_{jX}\|^2$. A Gaussian affinity kernel is then formed as:
\begin{align}\label{rbf}
    P = \exp(-\beta D^2),
\end{align}
with scale parameter $\beta>0$. DMAP replaces this raw affinity by the row-stochastic Markov-operator:
\begin{align*}
    P^+ = \diag(P\mathbf{1})^{-1}P,
\end{align*}
which defines a random walk on the sampled data cloud. Since $P^+$ is generally non-symmetric, it is convenient to instead diagonalize the similar symmetric-matrix $S = \diag(P\mathbf{1})^{-1/2}P\diag(P\mathbf{1})^{-1/2},$
which has the same eigenvalues as $P^+$ and yields the same spectral content up to the usual change of basis. The associated random-walk Laplacian is
\begin{align}
    \Delta^+ = I - P^+,
\end{align}
such that if $P^+\phi_n=\lambda_n\phi_n$, then
\begin{align}
    \Delta^+\phi_n = \mu_n \phi_n,\qquad\mu_n = 1-\lambda_n.
\end{align}
Thus the diffusion-map spectrum $\lambda_n\in[0,1]$ and the random-walk Laplacian spectrum $\mu_n\ge 0$ carry equivalent information, expressed either as slow Markov modes or as Laplacian decay rates.

The role of the kernel scale $\beta$ is subtle, but important. In the flat-kernel limit $\beta\to 0$, $P$ approaches the rank-one matrix $\mathbf{1}\mathbf{1}^\top$ and the Markov chain becomes trivial, dominated by a single stationary mode. In the opposite limit $\beta\to\infty$, the off-diagonal affinities vanish and $P$ approaches the identity, so the points become effectively disconnected and the spectrum spreads toward its full discrete rank \cite{barthelme2019_kernelFlatLimitSpectral}. 

For any finite set of distinct samples and any fixed $\beta>0$, the Gaussian kernel matrix, eq. \ref{rbf}, is generically strictly positive definite, and hence has full algebraic rank. The relevant notion is therefore not exact rank but \emph{effective} rank, denoted $r_{\mathrm{eff}}(\beta)$, which measures how many eigenmodes carry appreciable weight at scale $\beta$; this may be quantified, for example, by spectral thresholding, stable rank, or entropy rank \cite{hofmann2008_kernelMethods, buhmann2000_rbf}. From this perspective, $\beta$ does more than determine a neighborhood size: it controls the spectral resolution of the kernel, ranging from a coarse low-rank description to an increasingly localized representation of the data.

A natural heuristic for the dependence of $r_{\mathrm{eff}}(\beta)$ on scale comes from viewing the Gaussian kernel as a discrete heat operator. One then expects that modes with Laplace--Beltrami eigenvalues $\lambda_\ell \lesssim \beta$ remain appreciable at scale $\beta$. By Weyl's law \cite{chitour2024_weylSingularManifolds}, the number of such modes on a $d$-dimensional manifold scales as $\beta^{d/2}$, leading to the asymptotic estimate:
\begin{align}
r_{\mathrm{eff}}(\beta)\sim \beta^{d/2},
\end{align}
up to a geometry-dependent pre-factor and saturation at the finite sample size $N$.

This observation already hints at a limitation of interpreting DMAP as dimensionality reduction in the same sense as a geometric embedding. The classical inverse-spectral question ``Can one hear the shape of a drum?'' asks whether the spectrum of the Laplacian uniquely determines the underlying geometry; in general, it does not, since distinct non-isometric domains may be isospectral \cite{kac1966_hearShapeDrum, gordon1992_youCantHearShapeDrum}. The diffusion-map spectrum inherits this tension. It is highly informative about diffusion geometry and time scales, but eigenvalues alone do not determine a unique chart, parametrization, or ambient reconstruction. Recovering an embedding from spectral data is therefore an additional inverse problem, not something guaranteed by the spectrum itself. Here the relevant notion of \emph{isometry} is the preservation of intrinsic distances on $\mathscr{M}$: two embeddings are isometric if they induce the same Riemannian metric, even if they appear very different in ambient space. DMAP is designed to reflect intrinsic geometry, but this is precisely why a low-frequency spectral description need not coincide with a low-dimensional Euclidean chart. In short, one may hear aspects of diffusion geometry, but one does not in general obtain an embedding for free.

Although a low-dimensional embedding may often be approximated as a linear combination of diffusion eigenfunctions, this fact is only diagnostic when a target embedding is already available. In the unsupervised setting, no such target chart is given. The diffusion spectrum provides a basis of smooth modes, but it does not by itself specify which combination of these modes should be interpreted as a low-dimensional parametrization. Thus DMAP furnishes a representation of geometry, whereas dimensionality reduction requires an additional chart-selection principle, such as geodesic preservation, neighborhood preservation, reconstruction, or a task-specific criterion.

Recent work has clarified that diffusion-based constructions can support much richer geometric structure than a low-dimensional embedding alone. Spectral exterior calculus reconstructs differential forms and form Laplacians from the scalar Laplace--Beltrami spectrum \cite{berry2020_spectralExteriorCalculus}, while empirical Hodge-theoretic approaches build direct point-cloud analogues of exterior calculus under the manifold hypothesis \cite{lerch2025_empiricalHodgeLaplacians}. Classical discrete exterior calculus provides an important related simplicial precedent, though its mesh-based setting differs from the point-cloud and operator-learning perspective emphasized here \cite{desbrun2005_discreteExteriorCalculus}.

Interestingly, while the manifold hypothesis is a useful first approximation, the assumption of a single global intrinsic dimension is often too rigid for real data. Recent diffusion-geometric work makes this explicit: on a smooth manifold the pointwise metric has constant rank, but on empirical data the corresponding diffusion dimension can vary across space, suggesting that many datasets are better modeled as unions or stratifications of low-dimensional pieces than as a single smooth manifold \cite{jones2024_manifoldDiffusionGeometry,jones2026_diffusionGeometry}. Intrinsic dimension should therefore often be understood as a local, and possibly scale-dependent, quantity rather than a fixed global parameter. This perspective is consistent with recent work on stratification learning and on graph Laplacians near singularities \cite{aamari2024_theoryStratificationLearning, andersson2025_exploringSingularitiesGraphLaplacian}. It also clarifies the scope of different geometric constructions: Lerch and Wahl develop a higher-order empirical exterior calculus with convergence guarantees under the classical closed-submanifold model \cite{lerch2025_empiricalHodgeLaplacians}, whereas the broader diffusion-geometry program seeks to extend metric, differential, and topological constructions to variable-dimensional and singular data. Even outside pure geometry, this shift toward local structure is now operational in generative modeling, where Carré du champ flow matching replaces isotropic noise with anisotropic covariances adapted to local data geometry \cite{bamberger2025_cdcfm}. For the purposes of this paper, a constant intrinsic dimension should therefore be read as an idealized special case.

\section{Experiments}

We designed our experiments to distinguish between two different questions: whether a method captures the intrinsic geometry of the manifold, and whether it directly returns a useful low-dimensional chart. To make this distinction precise, we used a synthetic Swiss-roll for which the correct two-dimensional unrolling is known exactly. We then compared Isomap (IMAP), Diffusion Maps (DMAP), and UMAP across a range of latent dimensions, and evaluated how efficiently the ground-truth chart could be recovered from each representation.

\subsection{Ground-truth Swiss-roll construction}

\begin{figure}[t]
    \centering
    \includegraphics[width=0.7\linewidth]{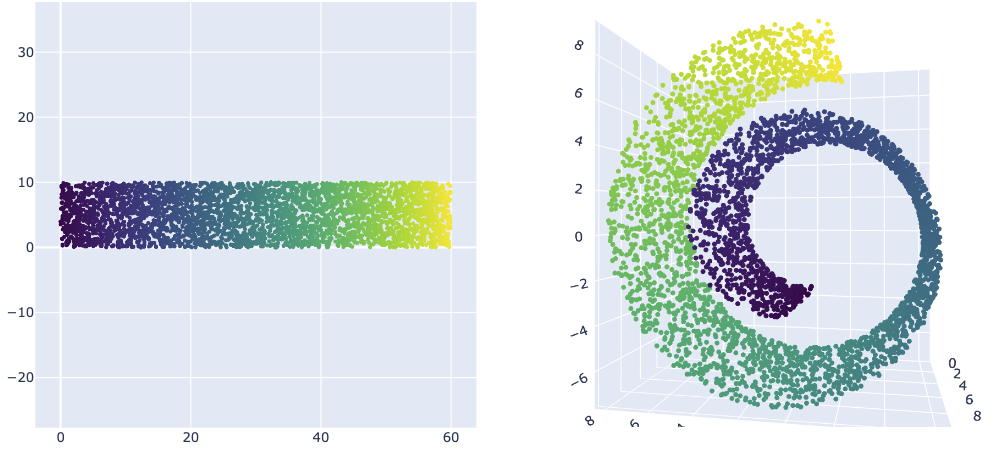}
    \caption{Above is a randomly sampled 2D sheet (left) and its perfect rolling in 3D (right).}
    \label{fig:swissroll}
\end{figure}

We generated a Swiss-roll dataset with known intrinsic coordinates by first sampling points uniformly from a rectangular sheet
\begin{align*}
(s,h)\in[0,W]\times[0,H],
\end{align*}
with $W=60$ and $H=10$. These sheet coordinates define the true isometric chart. We then embedded the sheet isometrically into $\mathbb{R}^3$ using the Archimedean Swiss-roll construction described earlier, producing ambient observations $X_i\in\mathbb{R}^3$ together with known intrinsic coordinates $Q_i\in\mathbb{R}^2$. Figure~\ref{fig:swissroll} illustrates the unrolled sheet, the rolled surface, and the inverse unrolling.

This construction removes the ambiguity present in standard manifold-learning benchmarks: the target chart is known exactly, so reconstruction error can be measured directly in the correct intrinsic coordinates.

\subsection{Representations and dimension scan}

For each dataset, we computed embeddings using IMAP, DMAP, and UMAP. For a target latent dimension $d$, each method produced a representation
\begin{align*}
U^{(d)}\in\mathbb{R}^{N\times d}.
\end{align*}
For IMAP and UMAP, the embedding was recomputed separately for each $d$. For DMAP, we instead computed a single ordered spectral basis up to $d_{\max}=1024$ and then truncated to the first $d$ coordinates as needed. This reflects the nested structure of diffusion coordinates: higher-dimensional trials simply extend the same spectral representation. Across all methods, we used the common scan $d\in\{1,2,3,4,5,6,7,8,16,32,64,128,256,512,1024\}$.

\subsection{Oracle linear readout}

Our goal was not to ask whether a method directly outputs the correct chart, but whether the correct chart is contained in the span of its learned coordinates. For each embedding $U^{(d)}$, we therefore fit the affine least-squares model
\begin{align*}
Q \approx U^{(d)}L + b,
\end{align*}
where $Q\in\mathbb{R}^{N\times 2}$ is the ground-truth unrolled sheet, $L\in\mathbb{R}^{d\times 2}$ is a linear readout matrix, and $b\in\mathbb{R}^2$ is a bias term. Equivalently, this can be implemented as a single least-squares solve using an augmented design matrix.

This readout should be interpreted as an oracle probe rather than an unsupervised algorithm. It answers the question: given a learned representation, how well can the correct chart be linearly recovered from it? In this way, the experiment separates representational capacity from chart selection. A method may contain the correct geometry without exposing it directly in its first two coordinates.

\subsection{Evaluation metrics}

For each method and each latent dimension $d$, we formed the reconstructed chart
\begin{align*}
\hat Q^{(d)} = U^{(d)}L + b
\end{align*}
and compared it with the ground truth. Our primary metric was the squared Frobenius error
\begin{align*}
\|Q-\hat Q^{(d)}\|_F^2,
\end{align*}
together with the corresponding mean-squared error and relative Frobenius error. Because the target chart is known exactly, these quantities directly measure how compactly each representation supports recovery of the isometric coordinates.

\subsection{Reconstruction results}

\begin{figure}[t]
    \centering
    \includegraphics[width=0.7\linewidth]{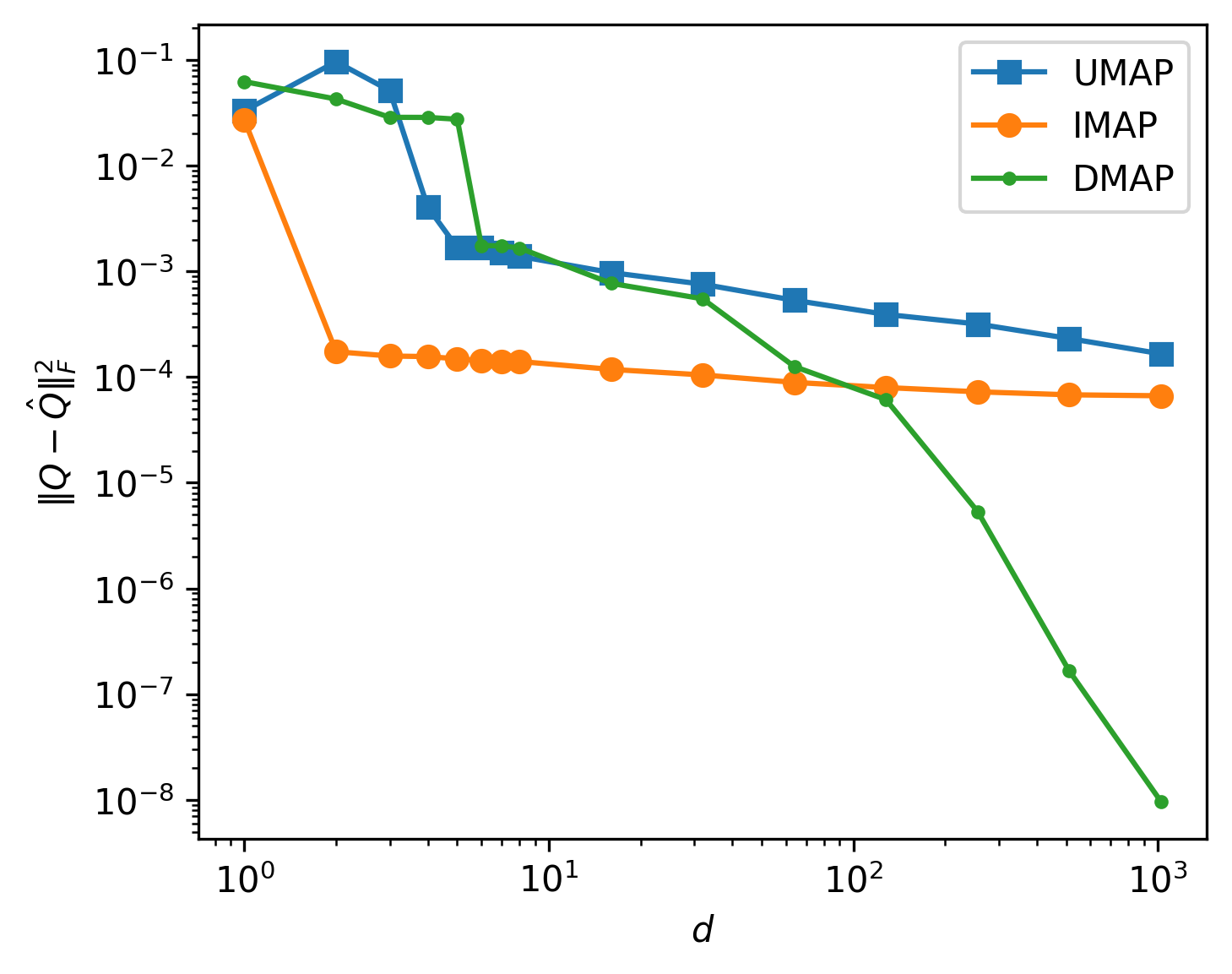}
    \caption{Reconstruction error of the ground-truth sheet as a function of latent dimension $d$. IMAP compresses the geometry most efficiently at low dimension, UMAP provides an intermediate tradeoff, and DMAP requires more modes but ultimately yields the most accurate reconstruction.}
    \label{fig:error}
\end{figure}

\begin{figure}[t]
    \centering
\begin{tikzpicture}
  \node[anchor=south west, inner sep=0] (img) at (0,0)
    {\includegraphics[width=0.9\linewidth]{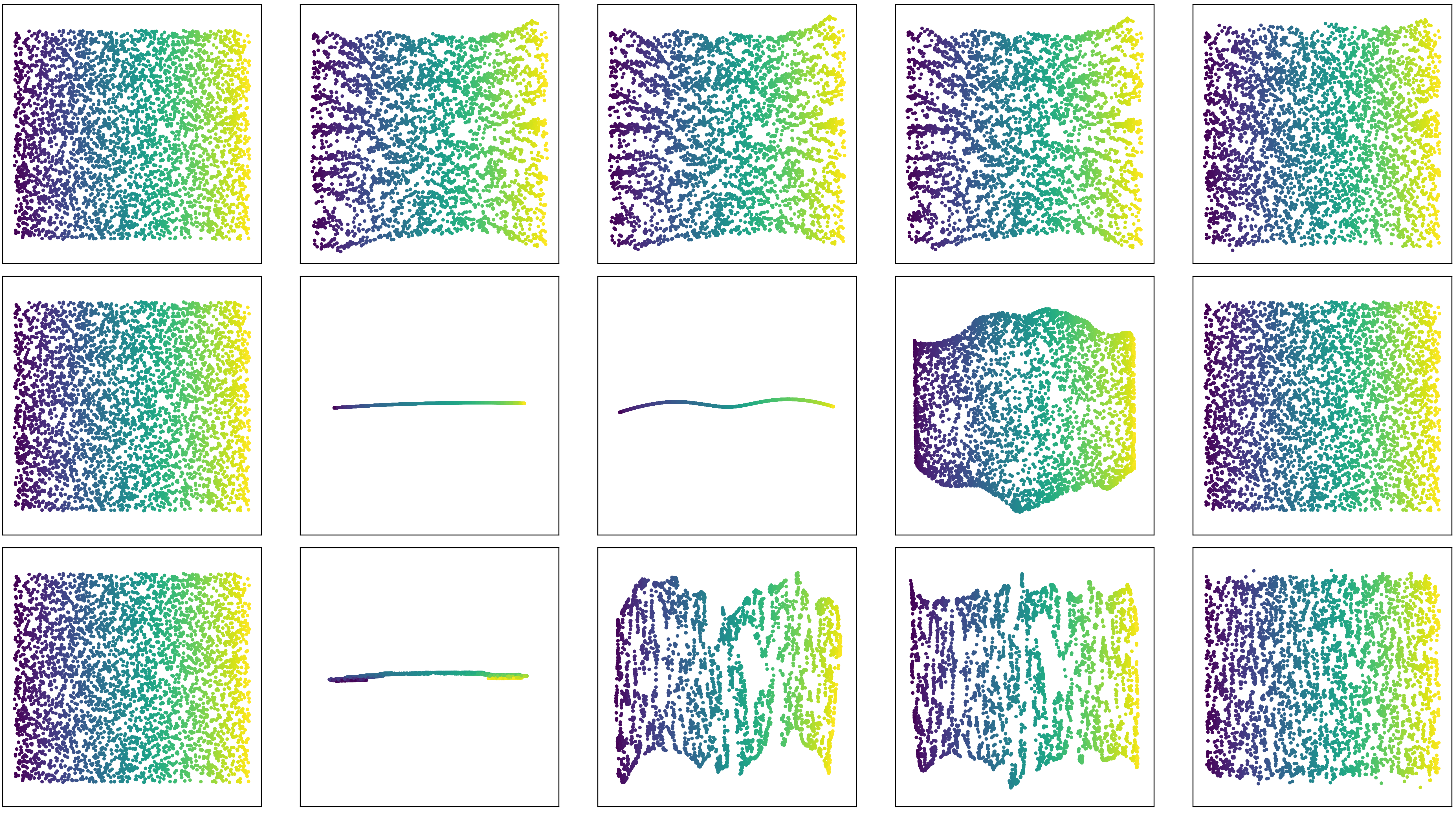}};
  \begin{scope}[x={(img.south east)},y={(img.north west)}]
    % coordinates now run from (0,0) bottom-left to (1,1) top-right
    %\node[fill=white, fill opacity=0.8, text opacity=1, rounded corners]
    %  at (0.20,1.00) {\small $d=2$};
    \node at (0.09,1.03) {\large truth};
    \node at (0.30,1.03) {\large $d=2$};
    \node at (0.50,1.03) {\large $d=4$};
    \node at (0.70,1.03) {\large $d=8$};
    \node at (0.90,1.03) {\large $d=1024$};
    %\draw[->, thick] (0.68,0.82) -- (0.61,0.73);
    \node at (-0.05,0.17) {\large UMAP};
    \node at (-0.05,0.50) {\large DMAP};
    \node at (-0.05,0.81) {\large IMAP};
  \end{scope}
\end{tikzpicture}
    \caption{Ground-truth sheet and affine reconstructions from IMAP, DMAP, and UMAP as the latent dimension $d$ increases. IMAP produces the correct sheet at small $d$, DMAP initially collapses to low-dimensional spectral modes before recovering the sheet at large $d$, and UMAP transitions between these behaviors.}
    \label{fig:IDU}
\end{figure}

Figure~\ref{fig:error} shows reconstruction error as a function of latent dimension. IMAP achieves low error at very small $d$, consistent with its explicit objective of recovering a compact isometric chart from approximate graph geodesics. UMAP exhibits intermediate behavior: substantial chart information is already present at low dimension, but reconstruction continues to improve as additional coordinates are included. DMAP behaves differently. Its first few coordinates are relatively inefficient for direct chart recovery, but the error decreases steadily with dimension and eventually falls below that of both IMAP and UMAP.

This difference reflects the distinct roles of the representations. IMAP exposes a usable chart quickly, but it is built from Dijkstra-type graph distance approximations, so its accuracy tends to saturate once the information in those approximate geodesics has been exhausted. DMAP, by contrast, does not directly solve the charting problem. Instead, it provides a spectral representation of the intrinsic geometry, with the relevant chart information distributed across many diffusion modes. As more modes are included, the oracle readout can exploit progressively richer geometric information, so the reconstruction improves monotonically and ultimately surpasses IMAP.

The qualitative reconstructions in Figure~\ref{fig:IDU} show the same pattern. IMAP recovers the sheet accurately at small $d$ but then changes little, UMAP improves more gradually, and DMAP begins with distorted low-dimensional projections yet recovers the unrolled sheet accurately once enough diffusion modes are combined. Thus, for DMAP, the failure of the first few coordinates to produce the correct chart does not mean that the geometry is absent; rather, the geometry is present in the diffusion basis but is not directly exposed as a leading low-dimensional embedding.

\subsection{Spectral structure of the DMAP readout}

\begin{figure}[t]
    \centering
    \includegraphics[width=0.7\linewidth]{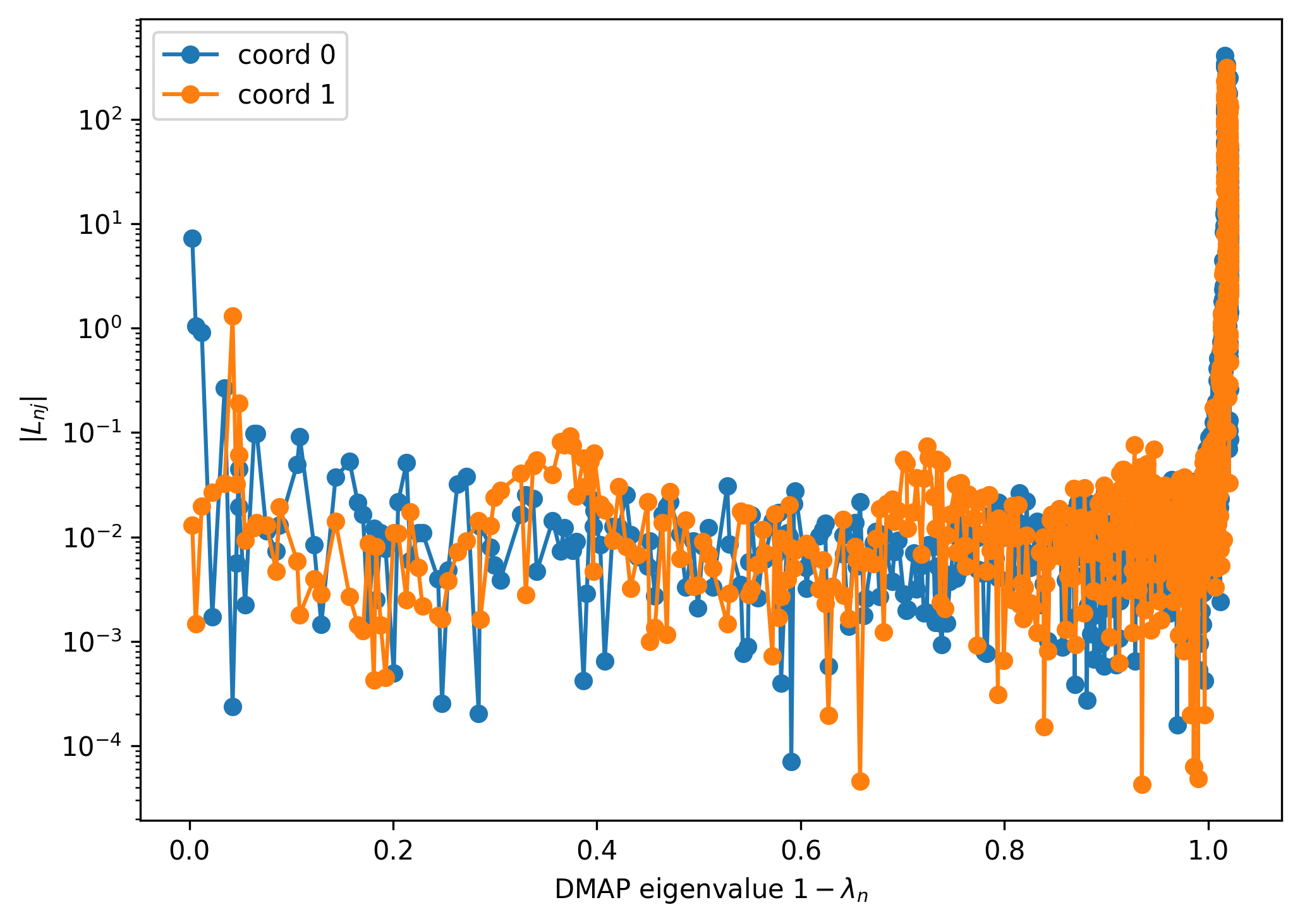}
    \caption{
    DMAP readout spectra for the two ground-truth coordinates, showing the coefficient magnitudes $|L_{ni}|$ versus the diffusion spectral variable $1-\lambda_n$. 
    Low values of $1-\lambda_n$ correspond to slow diffusion modes, while values near $1$ correspond to high-frequency modes. 
    The broad low-amplitude background and irregular spikes are consistent with residual least-squares gauge freedom and numerical noise, rather than a unique geometric signal.
    }
    \label{fig:dmap_spectra}
\end{figure}

To understand how the ground-truth chart is assembled from diffusion coordinates, we examined the affine readout matrix
\begin{align*}
L\in\mathbb{R}^{d\times 2},
\end{align*}
which assigns weights to diffusion modes when reconstructing the two intrinsic coordinates. Figure~\ref{fig:dmap_spectra} shows the magnitudes of these coefficients as a function of the diffusion spectral variable $1-\lambda_n$. The resulting spectra indicate that the target chart is distributed across multiple diffusion modes rather than being concentrated in a unique leading pair. Notably, Figure~\ref{fig:dmap_spectra} also exhibits isolated spikes at relatively high spectral frequency. These do not appear to signal a uniquely meaningful geometric coordinate; rather, they are more plausibly explained by residual gauge freedom in the least-squares readout together with numerical sensitivity in the high-frequency tail, i.e. the high frequency modes contribute nearly zero such that their coefficients are unimportant, hence their large values are by choice of the regressor.

\begin{figure}[t]
    \centering
    \includegraphics[width=1.0\linewidth]{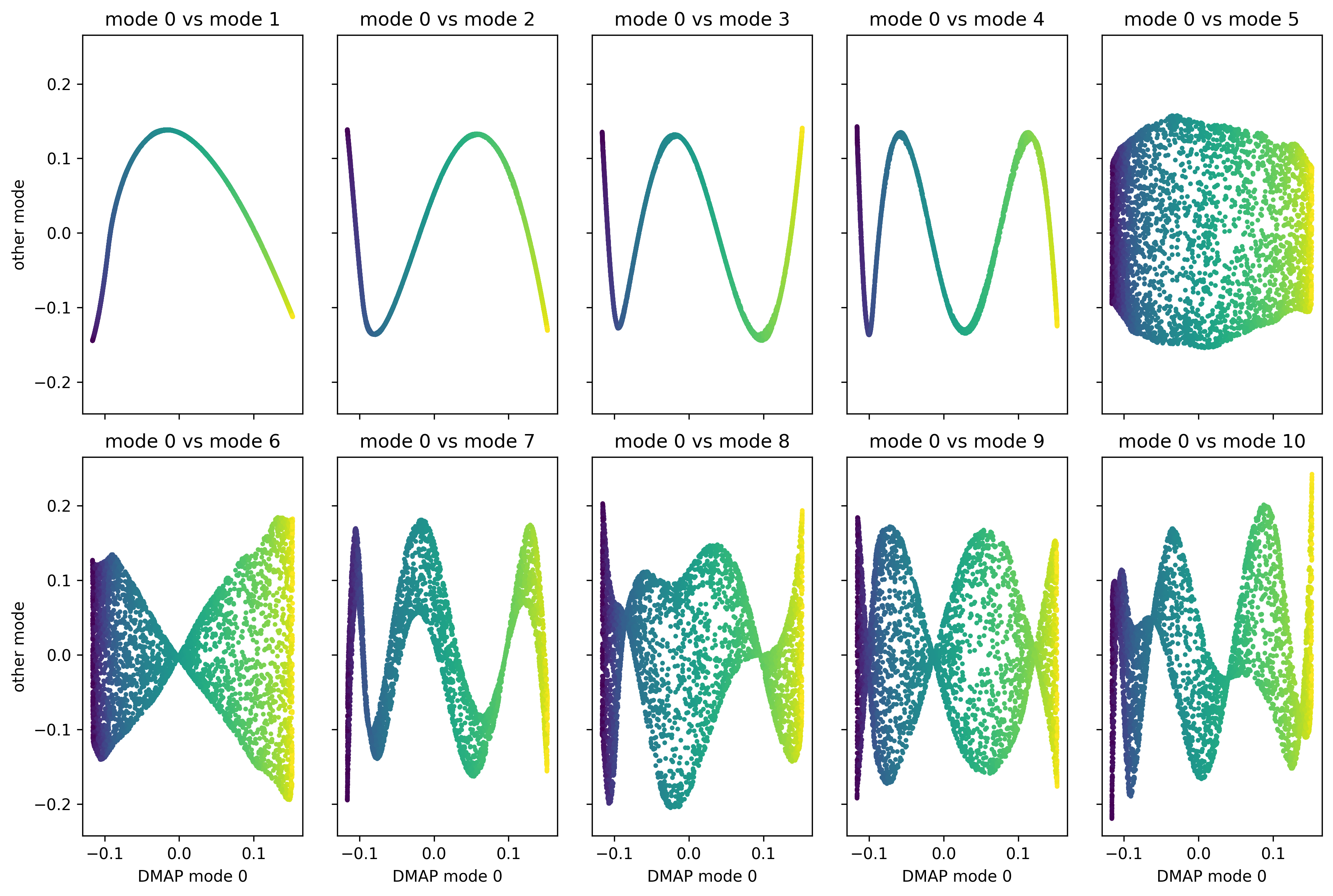}
    \caption{Two-dimensional charts formed by pairing DMAP mode 0 with modes 1--10. The figure shows that the unrolled sheet is not directly recovered from the first two diffusion modes; instead, sheet-like coordinates appear only for particular mode combinations, consistent with the spectrum in fig. \ref{fig:dmap_spectra} (mode 5), inline with the view of DMAP as a spectral basis rather than a canonical chart.}
    \label{fig:sheets}
\end{figure}

Figure~\ref{fig:sheets} makes this point visually by pairing DMAP mode $0$ with higher modes. The lowest pairings remain essentially curve-like, indicating redundancy with mode $0$, whereas certain higher pairings occupy substantial two-dimensional area and begin to resemble sheet-like parameterizations. This suggests a practical chart-selection heuristic: modes that increase the apparent local dimension of the joint embedding are plausible candidates for intrinsic coordinates. However, this criterion is only heuristic. Spectral orthogonality alone does not guarantee that the resulting map is injective or low-distortion, so such selections should ultimately be assessed through a local Jacobian or reconstruction criterion.

\section{Conclusion}

The main point of this paper is not that DMAP fail to capture meaningful low-dimensional structure, but that their role should be stated precisely. In the Swiss-roll example, the correct isometric chart can be recovered accurately from diffusion coordinates, but only after an external readout selects an appropriate combination of modes. DMAP therefore provide a powerful spectral representation of intrinsic geometry, but not by themselves a complete chart-selection principle.

Our experiments clarify this distinction. IMAP is the most efficient method when the objective is direct recovery of a low-dimensional isometric chart, while UMAP provides a compact embedding that supports good reconstruction at moderate dimension. DMAP behaves differently: its leading coordinates do not directly yield the desired unrolling, yet the correct chart lies in the span of its diffusion modes and can be recovered accurately through an appropriate linear combination. In this sense, DMAP is best understood as supplying the spectral ingredients for meaningful reduced coordinates, rather than uniquely determining those coordinates on their own.

This distinction matters in applications such as protein conformational dynamics and cryo-EM heterogeneity analysis, where diffusion coordinates often reveal slow collective structure in extremely high-dimensional data. Their practical value is therefore undeniable. But the scientifically relevant variable is often not a single diffusion coordinate; rather, it is a task-dependent combination of modes. What remains, then, is an inverse problem: how to pass from a diffusion basis to the physically or geometrically correct parametrization.

\newpage
\bibliographystyle{apalike}
\bibliography{refs}

\end{document}